\documentclass[10pt,twocolumn,letterpaper]{article}

\usepackage{cvpr}              

\usepackage{graphicx}
\usepackage{amsmath}
\usepackage{amssymb}
\usepackage{booktabs}

\usepackage{subcaption}
\usepackage{multirow}
\usepackage{hhline}
\usepackage{enumitem}
\usepackage[T1]{fontenc}
\usepackage{caption}

\usepackage[pagebackref,breaklinks,colorlinks]{hyperref}

\newlength\savewidth\newcommand\shline{\noalign{\global\savewidth\arrayrulewidth
\global\arrayrulewidth 1pt}\hline\noalign{\global\arrayrulewidth\savewidth}}

\usepackage[capitalize]{cleveref}
\crefname{section}{Sec.}{Secs.}
\Crefname{section}{Section}{Sections}
\Crefname{table}{Table}{Tables}
\crefname{table}{Tab.}{Tabs.}

\usepackage[accsupp]{axessibility}  

\def\cvprPaperID{8681} 
\def\confName{CVPR}
\def\confYear{2022}

\begin{document}

\title{Stand-Alone Inter-Frame Attention in Video Models}


\author{Fuchen Long$^{\dag}$, Zhaofan Qiu$^{\dag}$, Yingwei Pan$^{\dag}$, Ting Yao$^{\dag}$, Jiebo Luo$^{\S}$ and Tao Mei$^{\dag}$ \\
	$^{\dag}$JD Explore Academy, Beijing, China \\
	$^{\S}$University of Rochester, Rochester, NY USA \\
	{\tt\small\{longfc.ustc, zhaofanqiu, panyw.ustc, tingyao.ustc\}@gmail.com} \\
	{\tt\small\ jluo@cs.rochester.edu; tmei@jd.com} \\
}

\maketitle

\begin{abstract}
Motion, as the uniqueness of a video, has been critical to the development of video understanding models. Modern deep learning models leverage motion by either executing spatio-temporal 3D convolutions, factorizing 3D convolutions into spatial and temporal convolutions separately, or computing self-attention along temporal dimension. The implicit assumption behind such successes is that the feature maps across consecutive frames can be nicely aggregated. Nevertheless, the assumption may not always hold especially for the regions with large deformation. In this paper, we present a new recipe of inter-frame attention block, namely Stand-alone Inter-Frame Attention (SIFA), that novelly delves into the deformation across frames to estimate local self-attention on each spatial location. Technically, SIFA remoulds the deformable design via re-scaling the offset predictions by the difference between two frames. Taking each spatial location in the current frame as the query, the locally deformable neighbors in the next frame are regarded as the keys/values. Then, SIFA measures the similarity between query and keys as stand-alone attention to weighted average the values for temporal aggregation. We further plug SIFA block into ConvNets and Vision Transformer, respectively, to devise SIFA-Net and SIFA-Transformer. Extensive experiments conducted on four video datasets demonstrate the superiority of SIFA-Net and SIFA-Transformer as stronger backbones. More remarkably, SIFA-Transformer achieves an accuracy of 83.1\% on Kinetics-400 dataset. Source code is available at \url{https://github.com/FuchenUSTC/SIFA}.
\end{abstract}

\section{Introduction}

\begin{figure}[!tb]
      \vspace{-0.28in}
         \centering\includegraphics[width=0.42\textwidth]{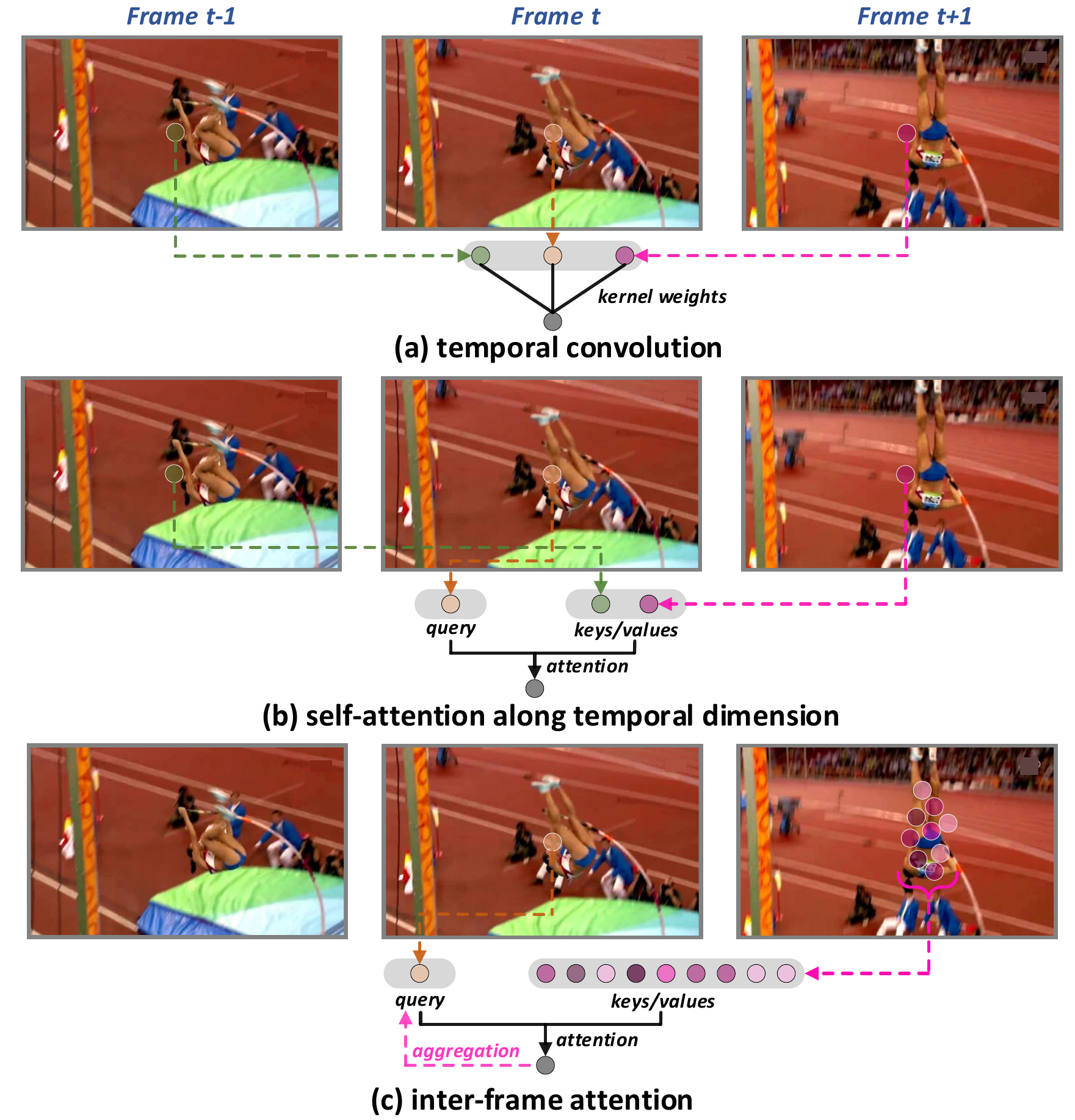}
         \vspace{-0.15in}
         \caption{\small Illustration of (a) temporal convolution, (b) self-attention along temporal dimension, and (c) our inter-frame~attention.}
         \label{fig1:1}
   \vspace{-0.30in}
\end{figure}

Video is an electronic representation of moving visual images and naturally forms the motion, which signifies a continuous change in position of objects or persons with time. Modeling such temporal dynamics is essential to the extension from understanding still images to videos. The recent advances generally suggest to leverage motion along two directions. One involves utilization of temporal convolutions by being integrated into space-time 3D convolutions \cite{Ji:PAMI,Tran:ICCV15} or explicitly co-working with spatial convolutions \cite{Tran:CVPR18,Carreira:CVPR17,Xie:ECCV18}. The other measures self-attention of each location over the temporal neighbors at the same spatial position across frames. Figure \ref{fig1:1}(a) and (b) conceptually depict the implementation of temporal convolution and self-attention along temporal dimension, respectively. The underlying spirit behind these operations originates from the foundation that the feature maps across frames should be well aligned. This assumption nevertheless may not always be valid in practice. Taking the three consecutive frames in Figure \ref{fig1:1} as an example, the same positions across frames highlighted in the circles correspond to different objects (person and track in the case) due to the motion of the athlete in pole vault. As such, performing temporal convolution or computing attention over these positions might be suboptimal for temporal feature aggregation.

To alleviate this issue, we propose to take the changes in video content caused by motion into account to enhance the alignment of feature maps across frames and eventually improve temporal aggregation. Technically, we develop inter-frame attention as shown in Figure \ref{fig1:1}(c) to characterize richer inter-frame correlation within a local neighboring region rather than only the same spatial location in consecutive frames. By doing so, inter-frame attention, on one hand, is beneficial more with large receptive fields, and on the other, manifests the emphasis of each location in the region to better achieve feature alignment. In an effort to nicely support the regions with large deformation, we further capitalize on the deformable design and estimate the offset to each spatial location. Moreover, we uniquely exploit the motion cues across frames to act as motion supervisory signal and re-scale the deformable feature re-sampling.

By delving into the deformation across frames to infer temporal attention within locally deformable region for temporal modeling, we present a novel Stand-alone Inter-Frame Attention (SIFA) block in video models. Specifically, we take each spatial location in the current frame as the query, and its temporal neighbors within the local region of the next frame are treated as keys/values accordingly to trigger the inter-frame attention learning. Note that in view of the irregular geometric transformations of objects, we sample the keys/values of temporal neighbors in a spatial deformation, which is learnt with additional guidance of the motion cues across frames. After that, SIFA block regards the estimated inter-frame attention of each temporal neighbor as its temporal correlation against query. Finally, we aggregate all temporal neighbors of nearby frames with inter-frame attention weights to further strengthen the query feature in current frame via temporal aggregation.

The SIFA block can be viewed as a stand-alone attention primitive for temporal modeling, and is readily pluggable to any 2D CNN or Vision Transformer backbones for video representation learning. By directly inserting SIFA block in ResNet \cite{Kaiming:CVPR16} and Swin Transformer \cite{Swin-ViT}, we construct two new video backbones, named as SIFA-Net and SIFA-Transformer, respectively. Through extensive experiments on a series of action recognition benchmarks, we demonstrate that our SIFA-Net and SIFA-Transformer outperform several state-of-the-art video backbones.

\section{Related Work}

We categorize existing research for video representation learning into hand-crafted and deep model based methods.

\textbf{Hand-crafted Representation}.
The early hand-crafted video feature techniques first detect spatio-temporal interest points and then describe them with local representations, such as STIP \cite{Laptev:IJCV05}, Histogram of Gradient and Histogram of Optical Flow \cite{Laptev:CVPR08}, 3D Histogram of Gradient \cite{Klaser:BMVC08}, and SIFT-3D \cite{Scovanner:MM07}.
Besides, Wang \emph{et al.} design the dense trajectory feature \cite{Wang:CVPR11} that samples dense local patches from each frame at various scales and tracks them in an optical flow field to convey motion cues in temporal domain. Nevertheless, these hand-crafted features are not optimized, thereby hardly to be generalized across different video tasks.

\textbf{Deep Learning based Representation}.
This direction first emerges by directly applying 2D CNN over video frames for video representation learning. For instance, Karpathy \emph{et al.} stack frame-level CNN features in a fixed size of window and then leverage spatial convolution to learn video representation \cite{Sports1M}.
Later in \cite{Simonyan:NIPS14}, the two-stream model is devised by utilizing two 2D CNN separately on visual frames and stacked optical flows.
This technique is further extended by exploring the convolution fusion \cite{Feichtenhofer:CVPR16}, temporal segment networks \cite{Wang:ECCV16,Feichtenhofer:CVPR17,Yao:AAAI21} and convolutional encoding \cite{Diba:CVPR17}.
To capture the long-term temporal dependency which is commonly ignored in some two-stream networks, LSTM-based methods \cite{Yue-Hei:CVPR15,Srivastava:ICML15} are designed to model long-range temporal dynamics in videos.

The aforementioned approaches only treat video as a sequence of frames or optical flows, while leaving the pixel-level temporal evolution across consecutive frames unexploited. 3D CNN based video feature \cite{Tran:ICCV15} is thus proposed to alleviate this issue by employing 3D convolutional kernels over short clips.
Furthermore, the subsequent works \cite{Carreira:CVPR17,Qiu:ICCV17,Qiu:CVPR19,Xie:ECCV18,Zhao:NIPS18} show that factorizing 3D convolution into 2D spatial convolution and 1D temporal convolution leads to better results and presents good generalization ability on localization task \cite{Long:CVPR19,Dong:MM19,Dong:CVPR21,Long:ECCV20, Long:TMM20}. Most recently, inspired by the impressive performances of applying self-attention from NLP field \cite{Vaswani:NIPS17} into image feature learning \cite{ViT,Swin-ViT,Li:PAMI},
TimeSformer \cite{Bertasius:ICML21} performs self-attention along the temporal dimension and designs five variants for temporal modeling. Nevertheless, these methods equipped with temporal convolution or temporal self-attention still suffer from the robustness problem due to object deformation across~frames.

Our work belongs to deep model based techniques that model temporal dynamics through self-attention. Unlike TimeSformer \cite{Bertasius:ICML21} that measures self-attention of each
location solely over its temporal neighbors at the same spatial location, SIFA mechanism performs inter-frame attention within a local neighboring region with large receptive fields. Moreover, SIFA block goes beyond the measure of inter-frame self-attention within regular local region, and capitalizes on locally deformable neighbors to tackle the irregular object deformation issue in temporal modeling.

\begin{figure}[!tb]
      \vspace{-0.20in}
      \centering
      \includegraphics[width=0.46\textwidth]{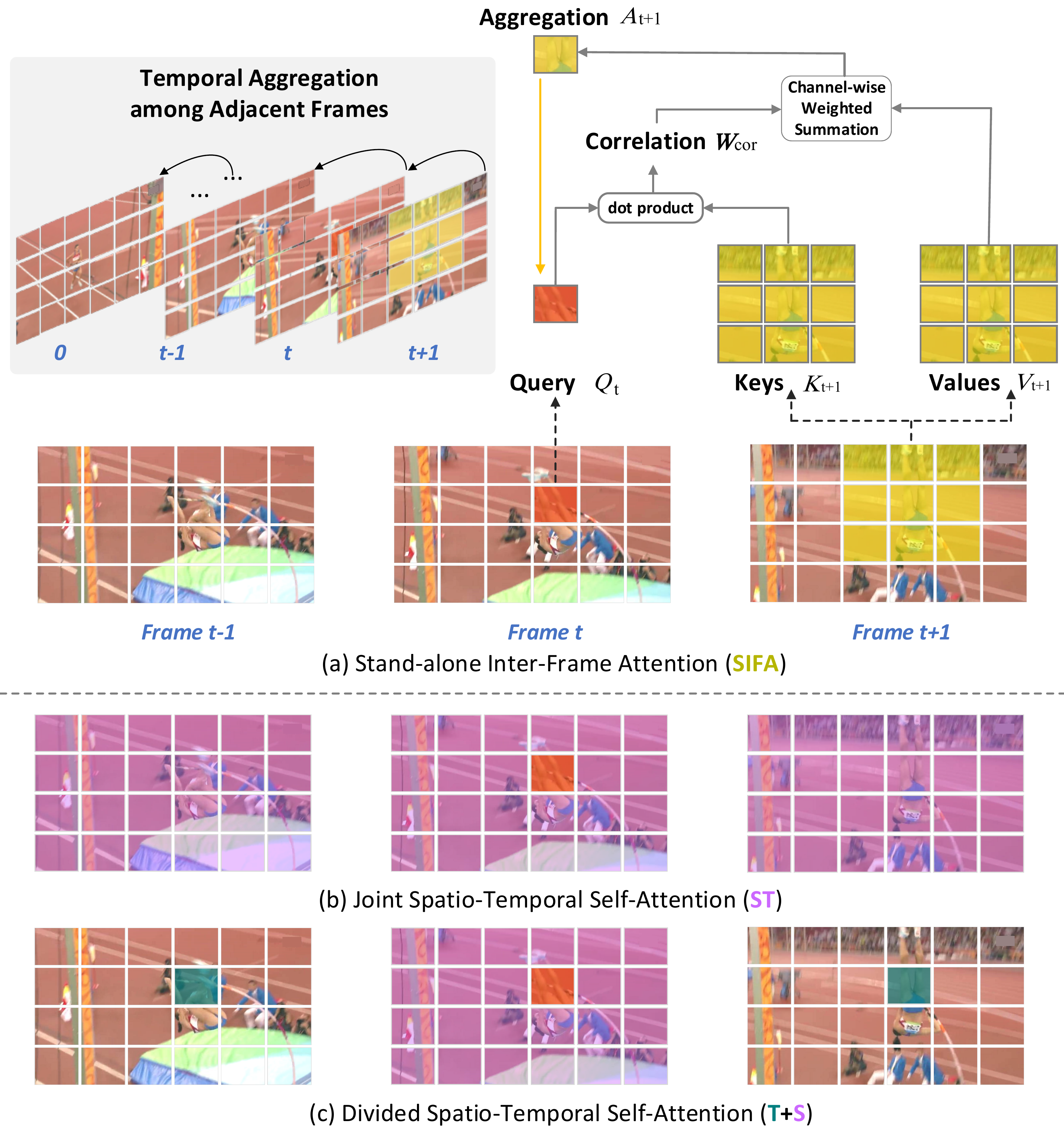}
      \vspace{-0.15in}
      \caption{\small Comparison between (a) our Stand-alone Inter-Frame Attention (\textbf{SIFA}) and two kinds of previous spatio-temporal attention, i.e., (b) joint spatio-temporal self-attention (\textbf{ST}) and (c) divided spatio-temporal self-attention (\textbf{T+S}). By visualizing each video clip as a sequence of frame-level patches, we denote in red the query patch and show its spatio-temporal neighbors in non-red colors for each attention mechanism. The patches without color are excluded for attention learning.
      Different from ST that employs attention over all frames holistically, T+S separately performs attention along the divided space and time dimensions.}
      \label{fig2:1}
      \vspace{-0.25in}
\end{figure}

\section{Our Approach}
We introduce a new Stand-alone Inter-Frame Attention (SIFA) for temporal modeling. SIFA exploits the temporal correlation within local region across consecutive frames, aiming to strengthen per-frame feature by aggregating its local neighbors in nearby frames via attention. Next, a novel stand-alone block in video models, i.e., SIFA block, is designed to perform such inter-frame attention over locally deformable region across frames. By plugging our SIFA block into 2D CNN (ResNet \cite{Kaiming:CVPR16}) and Vision Transformer (Swin Transformer \cite{Swin-ViT}), we further elaborate two video backbones, i.e., SIFA-Net and SIFA-Transformer.

\subsection{Stand-alone Inter-Frame Attention (SIFA)}
A natural way for temporal modeling in video representation learning is to use the 1D temporal convolution that conducts pixel-level feature aggregation across frames. However, this way solely captures motion clues among the same spatial locations along temporal dimension, while ignoring the inter-frame correlation at different spatial locations for temporal modeling. Inspired by the modeling of long-range dependencies via attention \cite{Vaswani:NIPS17,Wang:CVPR18}, we devise a new attention mechanism tailored for temporal modeling, i.e., Stand-alone Inter-Frame Attention (SIFA), that exploits the inter-frame correlation within local region for attention learning in an efficient manner. All the temporal neighbors within local region of nearby frames are aggregated with attention to strengthen per-frame feature.

Here we introduce the detailed formulation of our SIFA, as depicted in Figure \ref{fig2:1} (a).
Technically, let $F$ be the input 3D feature map with the size of $C\times L\times H\times W$, where $C$, $H\times W$, and $L$ denotes the channel size, spatial size, and temporal length, respectively. We first reshape $F$ into a 2D feature sequence $\{f_t\}_{t=0}^{L-1}$. Next, for $t$-th frame, we take its feature at the spatial location ($x, y$) as the query $Q_t\in \mathbb{R}^{C}$. Meanwhile, the features of ($t$+$1$)-th frame within the local region (size: $k\times k$ grid) centered at ($x,y$) are set as keys $K_{t+1}\in\mathbb{R}^{C\times \{k\times k\}}$ and values $V_{t+1}\in\mathbb{R}^{C\times \{k\times k\}}$. The correlation matrix $\mathbf{W}_{cor}$ between query $Q_t$ and keys $K_{t+1}$ is then calculated via dot production:
\begin{equation}\label{Eq2:1}
\mathbf{W}_{cor} = Q_{t} \odot K_{t+1},
\end{equation}
where $\odot$ denotes the matrix multiplication that measures the pairwise temporal correlation between query and its temporal neighbors (i.e., keys) within the local $k\times k$ grid.

Existing works commonly take the learnt correlation matrix $\mathbf{W}_{cor}\in\mathbb{R}^{1 \times \{k\times k\}}$ as pixel-level displacement information, and directly augment primary feature map with it to subserve flow estimation \cite{Philippe:ICCV13,Fischer:ICCV15}, geometric matching \cite{Rocco:CVPR17} and motion modeling \cite{Wang:CVPR20}. As an alternative, we capitalize on the correlation matrix as attention weights to dynamically aggregate the corresponding values within local region in nearby frame, targeting for enhancing query feature.
In particular, by taking the correlation matrix $\mathbf{W}_{cor}$ as the attention weights, the values $V_{t+1}$ within the local region are aggregated in a channel-wise manner:
\begin{equation}\label{Eq2:2}
A_{t+1} = \mathbf{W}_{cor} \odot [V_{t+1}]^T,
\end{equation}
where $A_{t+1}$ is the aggregated feature derived from the temporal neighbors of query, and the $[\cdot]^T$ denotes the matrix transpose. After that, we integrate the query with the aggregated feature, yielding the enhanced query feature $Y_t$ after temporal feature aggregation:
\begin{equation}\label{Eq2:3}
Y_t = Q_t + A_{t+1}.
\end{equation}

Accordingly, SIFA performs the inter-frame attention over each spatial location in $t$-th frame to mine its temporal correlation within local region of ($t$+$1$)-th frame. The feature map of each frame is thus strengthened by aggregating the features of local neighbors in the next frame via attention. In this way, we operate SIFA between every pair of adjacent frames in the input sequence. Note that for the last frame in the sequence, we conduct the inter-frame attention between this frame and itself, and enhance its feature map by itself through feature aggregation, thereby keeping the temporal length of output frame sequence as $L$.

\textbf{Connections with Previous Spatio-temporal Attention.}
Here we further discuss the detailed relations and differences between our SIFA and the previous spatio-temporal attention mechanisms. \cite{Bertasius:ICML21} introduces two kinds of spatio-temporal attention (i.e., joint or divided spatio-temporal self-attention) that employ self-attention over space and time for video representation learning. Specifically, the joint spatio-temporal self-attention (i.e., ST in Figure \ref{fig2:1} (b)) performs self-attention over the input features/patches of all frames holistically. The divided spatio-temporal self-attention (i.e., T+S in Figure \ref{fig2:1} (c)) separately applies the spatial attention within current frame and the temporal attention over the temporal neighbors in the same spatial location of nearby frames. Our SIFA also targets for exploring self-attention along temporal dimension for video modeling. Different from the global temporal attention over the holistic features/patches in ST, SIFA conducts the local temporal attention within local region across frames, which is computationally more efficient. Moreover, compared to S+T that only mines temporal evolution in the same spatial location of consecutive frames, SIFA captures the richer inter-frame correlation within local region for attention learning, thereby facilitating temporal modeling.

\subsection{SIFA Block}
Recall that our SIFA mechanism is devised to model the temporal evolution of objects within local region across consecutive frames. However, simply employing inter-frame attention over the equally-sized local region ($k\times k$ grid) inevitably ignores the irregular geometric transformations of objects in each frame, resulting in a sub-optimal solution. To alleviate this issue, we devise a SIFA block that applies inter-frame attention over the locally deformable region in nearby frames, which consists of the temporal neighbors sampled in a free-form spatial deformation.

\begin{figure}[!tb]
      \centering
      {\includegraphics[width=0.42\textwidth]{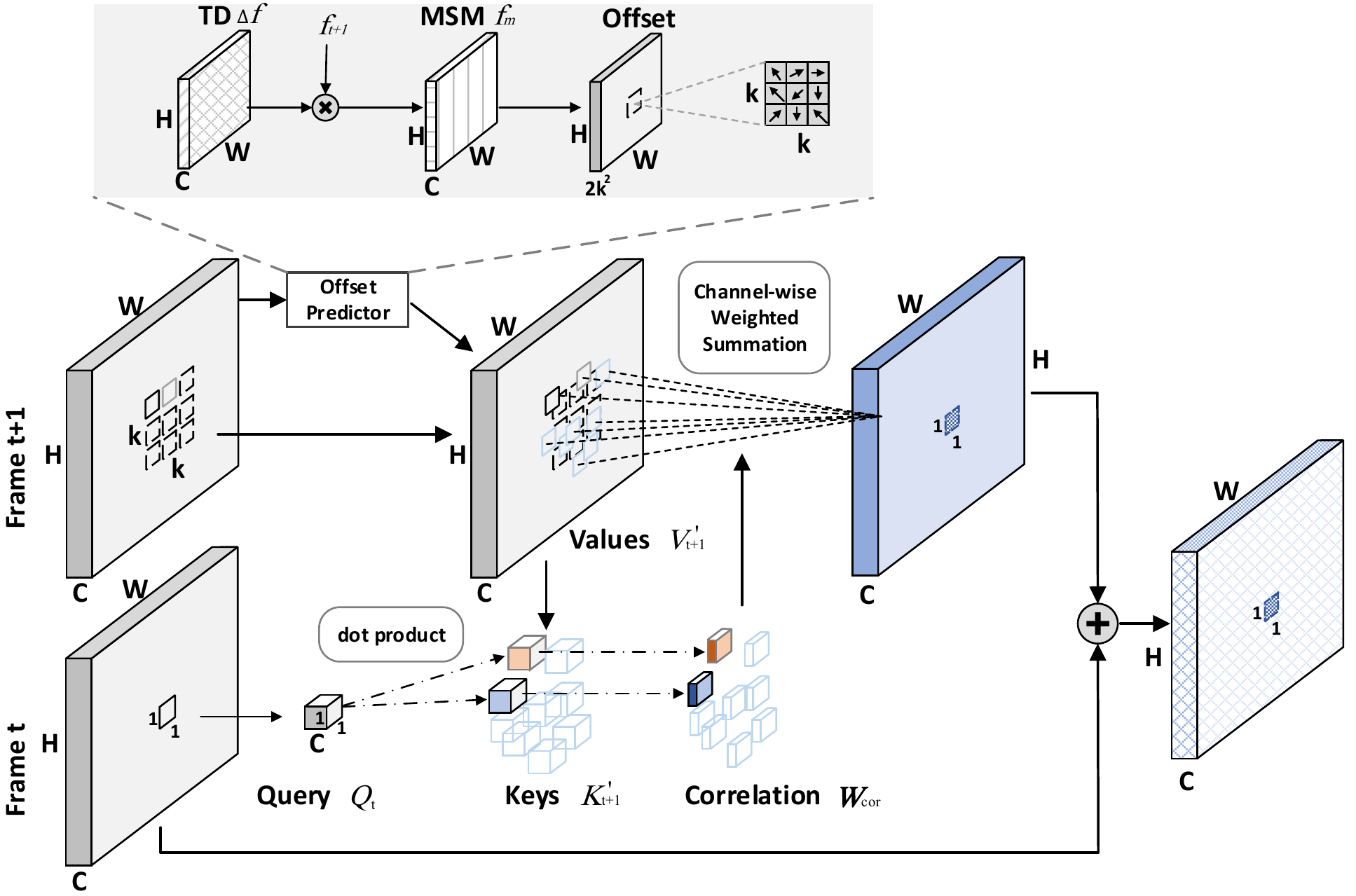}}
      \vspace{-0.16in}
      \caption{\small The detailed structure of our SIFA block.}
      \label{fig2:2}
      \vspace{-0.25in}
\end{figure}

The most typical way to operate deformable feature re-sampling is to augment the spatial sampling locations with additional offsets, that are predicted via a learnable offset estimator as in deformable ConvNets \cite{Dai:ICCV17}. Nevertheless, this offset estimator learns to infer the 2D offset of each spatial location solely based on the input feature map itself, while leaving the inherent motion clues across consecutive frames unexploited. Instead, we propose to estimate 2D offset of each spatial location within local region based on its motion saliency map (MSM), which acts as motion supervision to guide the deformable feature re-sampling. Figure \ref{fig2:2} shows the detailed structure of our SIFA block.

Formally, given each pair of consecutive frames (i.e., $t$-th frame ${f}_{t}$ and ($t$+$1$)-th frame $f_{t+1}$), we first compute the temporal difference (TD) in between:
\begin{equation}\label{Eq2:4}
\Delta{f} = f_{t+1} - {f}_{t}.
\end{equation}
Next, we employ a sigmoid operation over such temporal difference, leading to a normalized attention map. This attention map dynamically pinpoints the spatial locations in ($t$+$1$)-th frame that contain highly salient movements of objects. Therefore, the motion saliency map (MSM) $f_m$ is achieved by multiplying the feature map of ($t$+$1$)-th frame $f_{t+1}$ with the attention map:
\begin{equation}\label{Eq2:5}
{f}_{m} = sigmoid(\Delta f) * f_{t+1}.
\end{equation}
Conditioned on the motion saliency map $f_m$, we utilize an offset estimator to predict the 2D offset for each spatial location within the local region ($k\times k$ grid) of ($t$+$1$)-th frame $f_{t+1}$. Note that the offset estimator is implemented as a 2D convolutional layer with the output channel size of $2k^2$. More specifically, let ($\Delta a, \Delta b$) denote the estimated 2D offset of each spatial location $p=(a,b)$ within the $k\times k$ grid centered at the query location ($x,y$).
The corresponding irregular spatial location is thus represented as $p'=(a+\Delta a, b+\Delta b$).
Following \cite{Dai:ICCV17}, we sample the feature ${K'}_{t+1}(p')$ at each irregular spatial location $p'$ through bilinear interpolation:
\begin{equation}\label{Eq2:6}
{K'}_{t+1}(p') = \sum_{p}G(p,p')\cdot {K}_{t+1}(p),
\end{equation}
where $p'$ is the fractional spatial location and $p$ enumerates all integral spatial locations within the local region. ${K}_{t+1}(p)$ denotes the primary feature at regular spatial location $p$, and $G$ is bilinear interpolation kernel.
After sampling all the $k^2$ deformable features in ($t$+$1$)-th frame $f_{t+1}$, we take them as the keys ${K'}_{t+1}\in \mathbb{R}^{C\times\{k\times k\}}$ and values ${V'}_{t+1}\in \mathbb{R}^{C\times\{k\times k\}}$ with regard to the query $Q_t\in \mathbb{R}^{C}$ in $t$-th frame ${f}_{t}$. In this way, we perform SIFA mechanism over the locally deformable region in nearby frame, and further strengthen per-frame feature by aggregating these deformable features via attention:
\begin{equation}\label{Eq2:7}
\begin{split}
&\mathbf{W}_{cor} = Q_{t} \odot K'_{t+1}, \\
&A_{t+1} = \mathbf{W}_{cor} \odot [V'_{t+1}]^T,\\
&Y_{t} = Q_{t} + {A}_{t+1}.
\end{split}
\end{equation}
The enhanced feature $Y_{t}$ for $t$-th frame is finally taken as the output of SIFA block.

\subsection{2D CNN and Vision Transformer with SIFA}
\begin{figure}[!tb]
      \vspace{-0.1in}
      \centering
      {\includegraphics[width=0.38\textwidth]{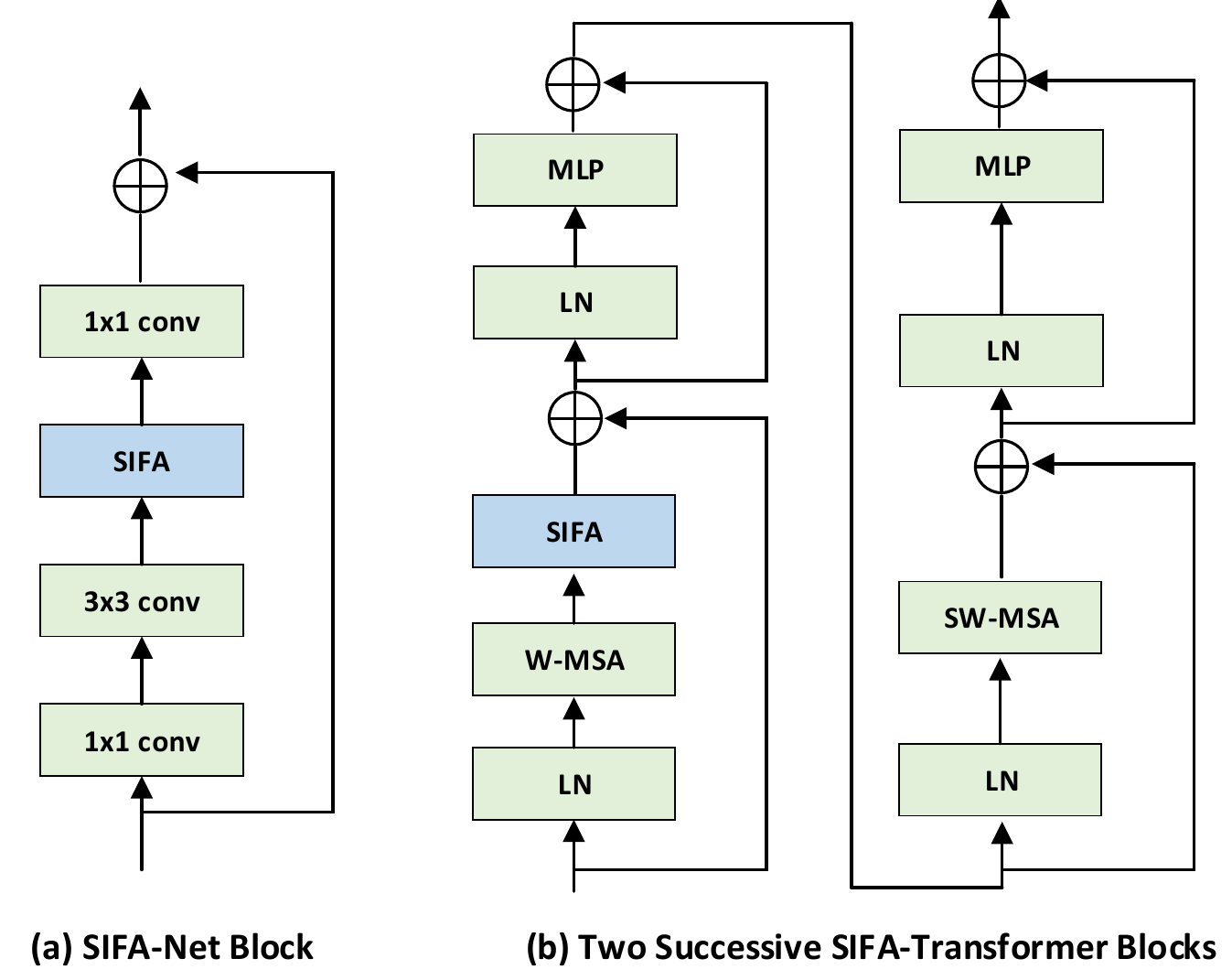}}
      \vspace{-0.14in}
      \caption{\small Basic blocks in (a) SIFA-Net and (b) SIFA-Transformer.}
      \label{fig2:3}
      \vspace{-0.27in}
\end{figure}

Our SIFA block acts as a stand-alone primitive for temporal modeling, and is pluggable to any 2D CNN or Vision Transformer architectures. Such design naturally upgrades these vision backbones with the capacity of temporal modeling, thereby boosting video representation learning. Here we present how to integrate SIFA block into existing 2D CNN (e.g., ResNet \cite{Kaiming:CVPR16}) and Vision Transformer (e.g., Swin Transformer \cite{Swin-ViT}). Figure \ref{fig2:3} depicts the two different constructions of equipping the basic building block in ResNet/Swin Transformer with our SIFA block, namely SIFA-Net and SIFA-Transformer, respectively.

\textbf{SIFA-Net.}
Most of existing video backbones \cite{Carreira:CVPR17,Tran:CVPR18,Xie:ECCV18,Qiu:ICML21} factorize the conventional 3D convolution into 2D spatial convolution and 1D temporal convolution, and the 1D temporal convolution is commonly plugged after the spatial convolutional layers of 2D CNN for temporal modeling across frames. We follow this typical paradigm and construct SIFA-Net by inserting SIFA block after the $3\times3$ convolution within each residual building block in ResNet \cite{Kaiming:CVPR16}. Note that we solely integrate the last three stages (i.e., $res_{3}$, $res_{4}$ and $res_{5}$) in ResNet with our SIFA block, thereby only increasing a small overhead to the computational cost. Finally, the global pooling is employed on the output feature to achieve the clip-level feature for video classification.

\textbf{SIFA-Transformer.}
Recently, computer vision field has witnessed the rise of Transformer-style architecture with self-attention \cite{ViT,Swin-ViT} in powerful vision backbones. Inspired by this, we further construct the Transformer-style video backbone, named as SIFA-Transformer, by integrating the Swin Transformer \cite{Swin-ViT} with our SIFA block. In particular, for every two successive Swin Transformer blocks in Swin Transformer, we directly insert the SIFA block after the MSA module with regular windowing configuration, leading to the two successive SIFA-Transformer blocks. Note that the output patch sequence of MSA module is reshaped into the sequence of feature map with the normal size ($C\times L\times H\times W$), which acts as the inputs of SIFA block. Based on the output reshaped sequence of feature map for the last block in SIFA-Transformer, we leverage the global pooling to obtain the clip-level feature.

\section{Experiments}
\subsection{Datasets and Implementation Details}
\textbf{Datasets.} We empirically evaluate the effectiveness of our SIFA-Net and SIFA-Transformer as video backbones on \textbf{Kinetics-400} \cite{Carreira:CVPR17}, \textbf{Kinetics-600} \cite{Kinetics:600}, \textbf{Something-Something V1 and V2} \cite{Goyal:SS} datasets.
The Kinetics-400 dataset consists of 300K videos derived from 400 action categories. Each video in Kinetics-400 is 10-second short clip cropped from the raw YouTube video.
In this dataset, all the 300K videos are divided into 240K, 20K, 40K for training, validation and testing, respectively.
The Kinetics-600 is an extended version of Kinetics-400, which includes around 480K videos from 600 action categories.
There are 390K, 30K, 60K clips in training, validation and testing sets, respectively.
In Something-Something V1 and V2 datasets, there are about 108K and 221K videos from 174 action categories, which are mostly for interaction-related recognition.
The training/validation/testing set includes 86K/11.5K/11K and 169K/25K/27K videos, respectively.

\begin{table*}
      \centering
      \vspace{-0.2in}
      \caption{\small Ablation study on SIFA block in SIFA-Net with 16-frame inputs on Kinetics-400 dataset. Top-1 and Top-5 accuracy (\%), and the computational cost (measured in GFLOPs) for forwarding one clip at inference are reported.}
      \vspace{-0.12in}
             \subcaptionbox{
             \label{tab1:sifa}
             \textbf{Stand-alone Inter-Frame Attention.}
             Comparisons among different variants of SIFA. All runs are constructed by plugging each block into $res_5$ stage of ResNet-50.}{
             \scalebox{0.80}[0.80]{
              \begin{tabular}{l|c|c c}
              \shline
              Model                         & GFLOPs &  Top-1  & Top-5  \\ \shline
              2D-ResNet                     & 23  &  72.0   & 90.3   \\
              \hline
              SIFA$_{C}$                    &  23  &  73.3   & 90.8   \\
              SIFA$_{R}$                    &  24  &  74.6   & 91.5   \\
              \hhline{*{4}{-}}
              SIFA                          &  24  &  75.4   &  92.9  \\
              \hhline{*{4}{-}}
              SIFA$^*$                      &  25  &  75.5   &  92.9 \\ \shline
              \end{tabular}
             }}
             \vspace{0.01in}
             \hspace{0.05in}
             \subcaptionbox{
             \label{tab1:offset}
             \textbf{Deformable Offset.} Comparisons across different ways on the measure of deformable offset in SIFA block. All runs are constructed by plugging each block into $res_5$ stage of ResNet-50.}{
             \scalebox{0.80}[0.80]{
              \begin{tabular}{l|c|c c}
              \shline
              Offset                      & GFLOPs &  Top-1  & Top-5  \\ \shline
              Regular (SIFA$_{R}$)        &  24  &  74.6   & 91.5   \\
              \hline
             Conv$_{2D}(f_{t+1})$        &  24   &  74.7   & 91.6   \\
             Conv$_{3D}(f)$              &  27   &  74.8   & 91.9   \\
             Conv$_{2D}(\Delta f)$       &  24   &  75.0   & 92.1   \\
             \hhline{*{4}{-}}
             Conv$_{2D}(f_m)$ (SIFA)     &  24   &  \textbf{75.4}   &  \textbf{92.9}  \\ \shline
              \end{tabular}
             }}
             \hspace{0.05in}
             \subcaptionbox{
             \label{tab1:region}
             \textbf{Local Region Size.} Comparisons by using different local region size $k$. All runs are constructed by plugging each block into $res_5$ stage of ResNet-50.}{
             \scalebox{0.80}[0.80]{
              \begin{tabular}{l|c|c c}
              \shline
              Size $k$            & GFLOPs &  Top-1  & Top-5  \\ \shline
              1 $\times$ 1                &  24  &  73.4   & 90.9   \\ \hline
              3 $\times$ 3                &  24  &  75.4   & 92.9   \\
              5 $\times$ 5                &  25  &  75.4   & 93.0   \\
              7 $\times$ 7                &  26  &  75.4   & 93.0   \\
              9 $\times$ 9                &  29  &  75.5   & 93.1   \\ \shline
              \end{tabular}
             }}
             \hspace{0.1in}
             \subcaptionbox{
             \label{tab1:stage}
             \textbf{Location of SIFA Block in SIFA-Net.} Effect of plugging SIFA block into different stages of ResNet-50.}{
             \scalebox{0.83}[0.83]{
             \begin{tabular}{c c c c|c|c c}
             \shline
              \multicolumn{4}{c|}{Stage} & \multicolumn{1}{|c|}{\multirow{2}{*}{\text{GFLOPs}}} &  \multicolumn{1}{|c}{\multirow{2}{*}{\text{Top-1}}}  &\multicolumn{1}{c}{\multirow{2}{*}{\text{Top-5}}}  \\ \hhline{*{4}{-}}
              $res_2$ & $res_3$ & $res_4$ & $res_5$ & \multicolumn{1}{|c|}{} & \multicolumn{1}{|c}{} & \multicolumn{1}{c}{} \\
              \shline
              \multicolumn{4}{c|}{}                  &  23  &  72.0   & 90.3   \\
              \hline
              &            &            & \checkmark &  24   &  75.4   & 92.9   \\
              &            & \checkmark & \checkmark &  24   &  76.2   & 93.0   \\
              & \checkmark & \checkmark & \checkmark &  25   &  \textbf{77.4}   & \textbf{93.3}  \\
              \checkmark & \checkmark & \checkmark & \checkmark &  26   &  77.4   & 93.2   \\ \shline
              \end{tabular}
             }}
             \hspace{0.4in}
             \subcaptionbox{
             \label{tab1:temporal}
             \textbf{Temporal Modeling.} Comparisons with different temporal modeling techniques (backbone: ResNet-50).}{
             \scalebox{0.83}[0.83]{
            \begin{tabular}{l|c|c c}
            \shline
            Temporal Modeling                       & GFLOPs&  Top-1  & Top-5  \\ \shline
            2D-ResNet                               &  23  &  72.0   & 90.3   \\
            \hline
            Temporal Conv \cite{Tran:CVPR18}        &  33  &  74.1   & 91.4   \\
            Temporal Shift \cite{JiLin:ICCV19}      &  23  &  74.7   & 91.4 \\
            Correlation \cite{Wang:CVPR20}          &  23  &  75.1   & 91.6   \\
            Temporal Difference \cite{Wang:CVPR21}  &  36  &  76.6   & 92.8   \\
            \hhline{*{4}{-}}
            SIFA                                    &  25   &  \textbf{77.4}   &  \textbf{93.3}  \\ \shline
            \end{tabular}
             }}  		

      \label{tab:ablations}
\vspace{-0.28in}
\end{table*}

\textbf{Network Training.} We implement our proposals on PyTorch framework. The mini-batch Stochastic Gradient Descent (SGD) algorithm with cosine learning rate \cite{Loshchilov:ICLR17} is employed for model optimization.
We fix the resolution of each frame as $224\times 224$, which is randomly cropped from the video clip resized with the short size in [$256,340$].
The input clip length is set in the range from $16$ to $64$.
We randomly flip each clip along horizontal direction for data augmentation, except for Something-Something V1 and V2 in view of the direction-related classes.
The size of local region $k$ in SIFA block is set as $3$.
We set the base learning rate as $0.04$ for SIFA-Net and $0.01$ for SIFA-Transformer. The dropout ratio is fixed as $0.5$.
The maximum training epoch number is $128$ in Kinetics datasets and $64$ in Something-Something datasets.
The mini-batch size is $256$ and the weight decay parameter is set as $0.0001$.

\textbf{Inference Strategy.} We adopt two kinds of inference strategies to evaluate SIFA-Net and SIFA-Transformer. For SIFA-Net, we follow the \textbf{3-crop} strategy as in \cite{Christoph:ICCV19} to crop three $256\times 256$ regions from each clip for evaluation. The video-level prediction score is thus achieved by averaging all scores from \textbf{10} uniform sampled clips. For SIFA-Transformer, we directly measure the video-level prediction score based on the \textbf{4} uniform sampled clips.

\subsection{Ablation Study on SIFA Block}
In this section, we perform a series of ablation studies to examine several technical choices of our proposed Stand-alone Inter-Frame Attention (SIFA) block in SIFA-Net.
Specifically, the deep architecture of SIFA-Net is constructed based on the backbone of ResNet-50, and we report the top-1 and top-5 accuracy on the validation set of Kinetics-400 for performance comparison.

\textbf{Stand-alone Inter-Frame Attention.}
We first investigate how each design in our SIFA block influences the overall performance of SIFA-Net. Table \ref{tab1:sifa} details the performance comparisons among different variants of SIFA block. Note that all ablated runs here are constructed by only plugging the SIFA variants into the building blocks at $res_5$ stage of ResNet-50. We start from a base block (\textbf{2D-ResNet}), which is a 2D CNN bottleneck block without any temporal modeling. By upgrading the base block with correlation operator \cite{Wang:CVPR20}, \textbf{SIFA$_{C}$} exhibits better performances, which show the merit of leveraging the pixel-wise movement information for temporal modeling. \textbf{SIFA$_{R}$} further aggregates its local temporal neighbors through inter-frame attention, leading to a performance boost of 74.6\% in top-1 accuracy. The results basically highlight the advantage of leveraging inter-frame attention to model the temporal correlation within local region across frames. \textbf{SIFA} is additionally benefited from the deformable feature re-sampling that explores the irregular geometric transformations of objects in the next frame, and the top-1 accuracy of SIFA finally achieves 75.4\%. In addition, we include an upgraded version of our SIFA block, i.e., \textbf{SIFA$^*$}, that aggregates the temporal neighbors within the locally deformable regions derived from both the previous and next frames, rather than solely involving the temporal neighbors from the next frame as in SIFA. Such temporal aggregation along both forward and backward directions in SIFA$^*$ only leads to a marginal performance improvement (0.1\% in top-1 accuracy), while requiring more GFLOPs.

\textbf{Deformable Offset.}
Next, we compare different approaches of predicting the 2D offset of each spatial location in nearby frame for deformable feature re-sampling in SIFA block. As mentioned in previous section, \textbf{SIFA$_{R}$} denotes the degraded version of SIFA and only employs inter-frame attention over regular local region in nearby frame, without deformable feature re-sampling. We also include three ablated runs of our SIFA, i.e., Conv$_{2D}$($f_{t+1}$), Conv$_{3D}$($f$), and Conv$_{2D}$($\Delta f$), that upgrade SIFA$_{R}$ with deformable feature re-sampling in multiple ways. Concretely, \textbf{Conv$_{2D}$($f_{t+1}$)} directly predicts the 2D offset solely based on the feature map of the next frame through 2D convolution.
\textbf{Conv$_{3D}$($f$)} leverages 3D convolution over the whole clip feature (i.e., the sequence of frame feature maps) to achieve the 2D offset of each spatial location within this clip.
\textbf{Conv$_{2D}$($\Delta f$)} exploits the temporal difference between adjacent frames to infer the 2D offset via 2D convolution.
Table \ref{tab1:offset} summarizes the performances across different ways on the measure of deformable offset.
In particular, by additionally exploring the spatial deformation of objects in each frame as in deformable ConvNets, Conv$_{2D}$($f_{t+1}$) slightly improves SIFA$_{R}$. The result basically validates the effectiveness of deformable feature re-sampling. Compared to Conv$_{2D}$($f_{t+1}$) that predicts the deformable offsets of each frame independently, Conv$_{3D}$($f$) jointly infers the offset of each spatial location based on the holistic frame sequence, and thus achieves better performances, while requiring more computational cost. Instead of using 3D convolution to capture motion clues for offset prediction in Conv$_{3D}$($f$), Conv$_{2D}$($\Delta f$) explicitly utilizes the temporal difference between consecutive frames to estimate 2D offset via 2D convolution, leading to performance improvements in an efficient way. Furthermore, by integrating the feature map of the next frame with the inter-frame motion saliency map for offset prediction, Conv$_{2D}(f_m)$ (i.e., our SIFA) obtains the highest performances.

\textbf{Local Region Size.}
To explore the effect of local region size $k$ for inter-frame attention learning in SIFA block, we evaluate the performance and computational cost by varying $k$ from 1 to 9 with an interval of 2 in Table \ref{tab1:region}. In the extreme case of $k=1$, only a single temporal neighbor at the same spatial location of nearby frame is taken as key to measure inter-frame attention. As such, the SIFA block degenerates to temporal convolution that only explores temporal evolution in the same spatial location across frames. With the use of larger local region size ($k=3$), the top-1 accuracy is significantly increased from 73.4\% to 75.4\%. That basically validates the merit of performing inter-frame attention over locally deformable region across consecutive frames. When further enlarging the local region size, the performances are less affected and meanwhile the computational cost is generally increased. Therefore, we empirically set the local region size $k$ as 3, which is seemingly to be a good trade-off between performance and computation~cost.

\begin{table}[!tb]
       \setlength{\belowcaptionskip}{-1pt}
       \centering
       \vspace{-0.20in}
       \caption{\small Performance comparisons on Kinetics-400. The input clip length of SIFA-Net is shown inside the bracket.}
       \vspace{-0.12in}
       \scalebox{0.71}[0.71]{
 \begin{tabular}{{l|l|c|c|c}}
        \shline
        \multicolumn{1}{c|}{\textbf{Approach}} & \textbf{Backbone} & \textbf{GFLOPs$\times$views} & \textbf{Top-1} & \textbf{Top-5} \\ \shline
        \multicolumn{5}{l}{\textbf{Convolutional Networks}} \\ \shline
        \text{I3D \cite{Carreira:CVPR17}}                       &  Inception & 108$\times$N/A  & 72.1  & 90.3 \\
        \text{TSN \cite{Wang:ECCV16}}                           &  Inception & 80$\times$10    & 72.5  & 90.2  \\
        \text{MF-Net \cite{Chen:ECCV18}}                        &  R34  & 11$\times$50         & 72.8  & 90.4 \\
        \text{R(2+1)D \cite{Tran:CVPR18}}                       &  R34  & 152$\times$10         & 74.3  & 91.4  \\
        \text{S3D \cite{Xie:ECCV18}}                            &  Inception & 71$\times$30    & 74.7  & 93.4  \\
        \text{TSM \cite{JiLin:ICCV19}}                          &  R50       & 33$\times$30    & 74.1  & 91.2   \\
        \text{TEINet \cite{Liu:AAAI20}}                         &  R50       & 33$\times$30    & 74.9  & 91.8   \\
        \text{TEA \cite{Yan:CVPR20}}                            &  R50       & 33$\times$30    & 75.0  & 91.8   \\
        \text{SlowFast \cite{Christoph:ICCV19}}                 &  R50+R50   & 36$\times$30    & 75.6  & 92.1  \\
        \text{NL I3D \cite{Wang:CVPR18}}                        &  R50       & 282$\times$30   & 76.5  & 92.6   \\
        \text{SmallBig \cite{Li:CVPR20}}                        &  R50       & 57$\times$30    & 76.3  & 92.5  \\
        \text{CorrNet \cite{Wang:CVPR20}}                       &  R50       & 115$\times$10   & 77.2  & -     \\
        \text{TDN \cite{Wang:CVPR21}}                           &  R50       & 72$\times$30    & 77.5  & 93.2  \\ \hline
        \text{SIFA-Net (16)}                                                       &  R50        & 25$\times$30    & {77.4}  & {93.3}    \\
              \text{SIFA-Net (32)}                                                        &  R50       & 51$\times$30    & {78.5}  & {93.6}    \\
              \text{SIFA-Net (64)}                                                        &  R50        &112$\times$30    & \textbf{80.1}  & \textbf{94.4}    \\ \hline \hline
        \text{ip-CSN \cite{Tran:ICCV19}}                        &  R101      & 83$\times$30    & 76.7    & 92.3   \\
        \text{SmallBig \cite{Li:CVPR20}}                        &  R101      & 418$\times$12   & 77.4    & 93.3   \\
        \text{NL I3D \cite{Wang:CVPR18}}                        &  R101      & 359$\times$30   & 77.7    & 93.3    \\
        \text{TDN \cite{Wang:CVPR21}}                           &  R101      & 132$\times$30   & 78.5    & 93.9   \\
        \text{CorrNet \cite{Wang:CVPR20}}                       &  R101      & 224$\times$30   & 79.2    &  -     \\
        \text{SlowFast \cite{Christoph:ICCV19}}                 &  R101+R101 & 234$\times$30   & 79.8    & 93.9   \\
        \hline
        \text{SIFA-Net (16)}                                     &  R101      &  39$\times$30  & {78.7}  & {94.0}   \\
        \text{SIFA-Net (32)}                                                        &  R101      &  78$\times$30  & {79.8}  & {94.2}   \\
        \text{SIFA-Net (64)}                                                        &  R101      &  157$\times$30 & \textbf{81.3}  & \textbf{95.2}   \\
        \shline
        \multicolumn{5}{l}{\textbf{Vision Transformer}} \\ \shline
        \text{TimeSformer \cite{Bertasius:ICML21}}              &  ViT-B     & 2,380$\times$3   & 80.7  & 94.7 \\
        \text{ViViT \cite{ViViT}}                               &  ViT-L     & 3,992$\times$12  & 81.3  & 94.7 \\
        \text{MViT \cite{Fan:MVIT}}                             &  MViT-B    & 455$\times$9    & 81.2  & 95.1 \\
        \text{Video-Swin \cite{Liu:V-Swin}}                     &  Swin-B    & 282$\times$12    & 82.7  & 95.5 \\
        \hline
        \text{SIFA-Transformer}                                 &  Swin-B    & 270$\times$12    & \textbf{83.1}  & \textbf{95.7} \\
        \shline
        \end{tabular}
       }
    \label{table5:2}
    \vspace{-0.3in}
\end{table}

\textbf{Location of SIFA Block in SIFA-Net.}
To show the relationship between performance and the location of SIFA block in SIFA-Net, we progressively plug SIFA blocks into the stages in ResNet-50 backbone, and compare the performances.
The results shown in Table \ref{tab1:stage} indicate that inserting SIFA blocks into more stages can generally improve the performances, while increasing the computation cost.
When taking a closer look at the top-1 and top-5 accuracy of different locations of SIFA block, the integration of SIFA blocks in the last three stages ($res_3$, $res_4$, and $res_5$) contributes more to the performance boosts. No significant performance improvement is attained when further plugging SIFA block into $res_2$ stage.
Accordingly, we solely integrate the last three stages in ResNet-50 with SIFA blocks, and seek a good accuracy-computation cost balance.

\textbf{Temporal Modeling.}
We also compare our SIFA with other existing temporal modeling techniques.
Table \ref{tab1:temporal} summarizes the results by integrating the ResNet-50 backbone with different temporal modeling blocks.
Overall, our SIFA exhibits better performances than other temporal modeling approaches with less or similar GFLOPs. The results generally indicate the advantage of exploring the deformation across frames to estimate local self-attention for temporal aggregation.
In particular, by explicitly capturing motion displacement across frames, Correlation \cite{Wang:CVPR20} outperforms Temporal Conv \cite{Tran:CVPR18}. Temporal Difference \cite{Wang:CVPR21} further boosts the performances by additionally modeling long-term motion. Nevertheless, the performances of Temporal Difference are still lower than that of our SIFA which exploits inter-frame attention for temporal modeling.

\begin{table}[!tb]
       \setlength{\belowcaptionskip}{-1pt}
       \centering
       \vspace{-0.20in}
       \caption{\small Performance comparisons on Kinetics-600. The input clip length of SIFA-Net is shown inside the bracket.}
       \vspace{-0.12in}
       \scalebox{0.71}[0.71]{
        \begin{tabular}{{l|l|c|c|c}}
        \shline
        \multicolumn{1}{c|}{\textbf{Approach}} & \textbf{Backbone} & \textbf{GFLOPs$\times$views} & \textbf{Top-1} & \textbf{Top-5} \\ \shline
        \multicolumn{5}{l}{\textbf{Convolutional Networks}} \\ \shline
        \text{I3D \cite{Carreira:CVPR17}}                       &  Inception & 108$\times$N/A   & 71.9  & 90.1 \\
        \text{SlowFast \cite{Christoph:ICCV19}}                 &  R50+R50   & 36$\times$30    & 78.8  & 94.0  \\ \hline
        \text{SIFA-Net (16)}							        &  R50	     & 25$\times$30    & {79.6}  & {94.5}    \\
		\text{SIFA-Net (32)}							        &  R50       & 51$\times$30    & {80.5}  & {95.2}    \\
		\text{SIFA-Net (64)}							        &  R50	     &112$\times$30    & \textbf{82.1}  & \textbf{95.8}    \\ \hline \hline
        \text{SlowFast \cite{Christoph:ICCV19}}                 &  R101+R101 &234$\times$30    & 81.8    & 95.1 \\
        \text{X3D-XL \cite{Feichtenhofer:CVPR20}}               &  custom    & 48$\times$30    & 81.9    & 95.5   \\ \hline
        \text{SIFA-Net (16)}                                    &  R101      &39$\times$30    & {80.8}  & {95.2}   \\
        \text{SIFA-Net (32)}							        &  R101      &78$\times$30    & {81.6}  & {95.5}   \\
        \text{SIFA-Net (64)}							        &  R101      &157$\times$30   & \textbf{83.2}  & \textbf{95.9}   \\
        \shline
        \multicolumn{5}{l}{\textbf{Vision Transformer}} \\ \shline
        \text{TimeSformer \cite{Bertasius:ICML21}}              &  ViT-B     & 1,703$\times$3   & 82.4  & 96.0 \\
        \text{ViViT \cite{ViViT}}                               &  ViT-L     & 3,992$\times$12  & 83.0  & 95.7 \\
        \text{MViT \cite{Fan:MVIT}}                             &  ViT-B     & 236$\times$5     & 83.8  & 96.3 \\
        \text{Video-Swin \cite{Liu:V-Swin}}                     &  Swin-B    & 282$\times$12    & 84.0  & 96.5 \\
        \hline
        \text{SIFA-Transformer}                                 &  Swin-B    & 270$\times$12    & \textbf{84.5}  & \textbf{96.9} \\
        \shline
        \end{tabular}
       }
    \label{table5:3}
    \vspace{-0.25in}
\end{table}

\subsection{Comparisons with State-of-the-Art Methods}
We compare SIFA-Net and SIFA-Transformer with various state-of-the-art techniques on Kinetics-400, Kinetics-600, and Something-Something V1 (SSv1) and V2 (SSv2) datasets. All runs are briefly grouped into two paradigms: Convolutional Networks and Vision Transformer.
Note that we implement SIFA-Net in two kinds of backbones, i.e., ResNet-50 (R50) and ResNet-101 (R101), and the input clip length is varied in the range of \{16, 32, 64\}.
The SIFA-Transformer is constructed based on the backbone of Swin Transformer (Swin-B) with the fixed input clip length (64 frames).
The computational cost is measured in GFLOPs $\times$ views, and the views represent the number of clips sampled from the full video at inference.

\begin{table}[!tb]
       \setlength{\belowcaptionskip}{-1pt}
       \centering
       \vspace{-0.22in}
       \caption{\small Performances on Something-Something V1 and V2. The input clip length of SIFA-Net is shown inside the bracket.}
       \vspace{-0.15in}
       \scalebox{0.70}[0.70]{
\begin{tabular}{{l|l|c|c c|c c}}
        \shline
        \multicolumn{1}{c|}{\multirow{2}{*}{\textbf{Approach}}} & \multirow{2}{*}{\textbf{Backbone}} & {\textbf{GFLOPs}} & \multicolumn{2}{c|}{\textbf{SSv1}}  & \multicolumn{2}{c}{\textbf{SSv2}} \\ \hhline{*{3}{~}*{4}{-}}
        &                                    &           \textbf{$\times$views}            &  \textbf{Top-1}  &  \textbf{Top-5}  &  \textbf{Top-1}  &  \textbf{Top-5} \\ \shline
        \multicolumn{7}{l}{\textbf{Convolutional Networks}} \\ \shline
        \text{NL I3D+GCN \cite{Wang:ECCV18}}                    &  R50       &  606           & 46.1 & 76.8 &  -   &  - \\
        \text{CPNet \cite{Xingyu:CVPR19}}                       &  R34       &  N/A           &  -   &  -   & 57.7 & 84.0 \\
        \text{TSM \cite{JiLin:ICCV19}}                          &  R50       &  98            & 47.2 & 77.1 & 63.4 & 88.5 \\
        \text{TAM \cite{Fan:NIPS19}}                            &  R50       &  48            & 48.4 & 78.8 & 61.7 & 88.1 \\
        \text{GST \cite{Luo:ICCV19}}                            &  R50       &  59            & 48.6 & 77.9 & 62.6 & 87.9 \\
        \text{SmallBig \cite{Li:CVPR20}}                        &  R50       & 105            & 49.3 & 79.5 & 62.3 & 88.5 \\
        \text{CorrNet \cite{Wang:CVPR20}}                       &  R50       & 115$\times$10  & 49.3 &  -   &  -   &  -   \\
        \text{ACTION-Net \cite{WangAct:CVPR21}}                 &  R50       & 69             &  -   &  -   & 64.0 & 89.3 \\
        \text{STM \cite{Jiang:ICCV19}}                          &  R50       & 67$\times$30   & 50.7 & 80.4 & 64.2 & 89.8  \\
        \text{MSNet \cite{Kwon:ECCV20}}                         &  R50       & 67             & 52.1 & 82.3 & 64.7 & 89.4 \\
        \text{TEINet \cite{Liu:AAAI20}}                         &  R50       & 99             & 52.5 &  -   & 65.5 & 89.8   \\
        \text{MG-TEA \cite{Zhi:ICCV21}}                         &  R50       &  N/A           & 53.2 &  -   & 63.8 &  -     \\
        \text{TDN \cite{Wang:CVPR21}}                           &  R50       & 72             & 53.9 & 82.1 & 65.3 & 89.5 \\ \hline
              \text{SIFA-Net (16)}                                                        &  R50        & 25$\times$3    & 52.7 & 81.9 & 64.8 & 89.4 \\
        \text{SIFA-Net (32)}                                                       &  R50        & 51$\times$3    & 54.0 & 82.2 & 66.0 & 89.6 \\
        \text{SIFA-Net (64)}                                                       &  R50        & 112$\times$3   & \textbf{55.2}  & \textbf{83.3}  & \textbf{66.9} & \textbf{90.7}  \\ \hline \hline
        \text{GSM \cite{Sudhakaran:CVPR20}}                     &  Inception & 268            & 55.2 & -   & -  &  - \\
        \text{CorrNet \cite{Wang:CVPR20}}                       &  R101      & 224$\times$30  & 53.3 & -   & -  &  - \\
        \text{MG-TEA \cite{Zhi:ICCV21}}                         &  R101      & N/A            & 53.3 & -   & 64.8 & - \\
        \text{TDN \cite{Wang:CVPR21}}                           &  R101      & 132            & 55.3 & 83.3 & 66.9 & 90.9 \\ \hline
        \text{SIFA-Net (16)}                                                       &  R101       & 39$\times$3    & 53.7 & 82.0 & 65.9 & 89.8 \\
        \text{SIFA-Net (32)}                                                       &  R101       & 78$\times$3    & 55.4 & 83.1 & 67.3 & 91.1 \\
        \text{SIFA-Net (64)}                                                       &  R101      & 157$\times$3   & \textbf{56.1} & \textbf{84.0}  & \textbf{68.1} & \textbf{92.0}   \\
        \shline
        \multicolumn{7}{l}{\textbf{Vision Transformer}} \\ \shline
        \text{TimeSformer \cite{Bertasius:ICML21}}              &  ViT-B     & 1,703$\times$3 &  -  & -  & 62.5 & - \\
        \text{ViViT \cite{ViViT}}                               &  ViT-L     & 903            &  -  & -  & 65.4 & 89.8 \\
        \text{MViT \cite{Fan:MVIT}}                             &  ViT-B     & 455$\times$3   &  -  & -  & 67.7 & 90.9 \\
        \text{Video-Swin \cite{Liu:V-Swin}}                     &  Swin-B    & 321$\times$3   &  -  & -  & 69.6 & 92.7 \\
        \hline
        \text{SIFA-Transformer}                                 &  Swin-B    & 270$\times$3   &  \textbf{57.3} & \textbf{85.1}  & \textbf{69.8}  & \textbf{93.1} \\
        \shline
        \end{tabular}
       }
    \label{table5:4}
    \vspace{-0.26in}
\end{table}

Table \ref{table5:2} summarizes the performance comparisons on Kinetics-400. For the group of Convolutional Networks, our SIFA-Net leads to better performances against other baselines.
In particular, SIFA-Net (32) in R50 backbone obtains 78.5\% top-1 accuracy, and outperforms the best competitor TDN by 1.0\% but with $\sim$30\% less computation cost in GFLOPs.
By sampling more frames in each clip for temporal modeling, SIFA-Net (64) improves the top-1 accuracy from 78.5\% to 80.1\%.
The superior results of SIFA-Net generally demonstrate the advantage of integrating 2D CNN with inter-frame attention to enable temporal modeling.
When further inserting SIFA block into a state-of-the-art 2D Vision Transformer backbone (Swin Transformer), SIFA-Transformer manages to achieve the best performance (83.1\% in top-1 accuracy) on Kinetics-400.
The performance of SIFA-Transformer is comparable to the superior 3D Vision Transformer (Video-Swin), but requires less computation cost.
The performance trends on Kinetics-600 are similar with those on Kinetics-400 as shown in Table \ref{table5:3}.
The results again verify the impact of SIFA block in both 2D CNN and Vision Transformer backbones for video representation learning.
Table \ref{table5:4} lists the performances on both SSv1 and SSv2 datasets.
Particularly, we follow the one-clip and 3-crop settings \cite{Bertasius:ICML21,Fan:MVIT,Liu:V-Swin} for testing on Something-Something.
Similarly, SIFA-Net (64) in R50 and R101 backbones surpasses the best competitor TDN by 1.3\%/1.6\% and 0.8\%/1.2\% in top-1 accuracy on SSv1/SSv2, respectively.
Furthermore, by plugging SIFA block into Swin-B backbone, our SIFA-Transformer obtains the best performances on both SSv1 and SSv2 datasets.

\begin{figure}[!tb]
\vspace{-0.22in}
\centering\includegraphics[width=0.45\textwidth]{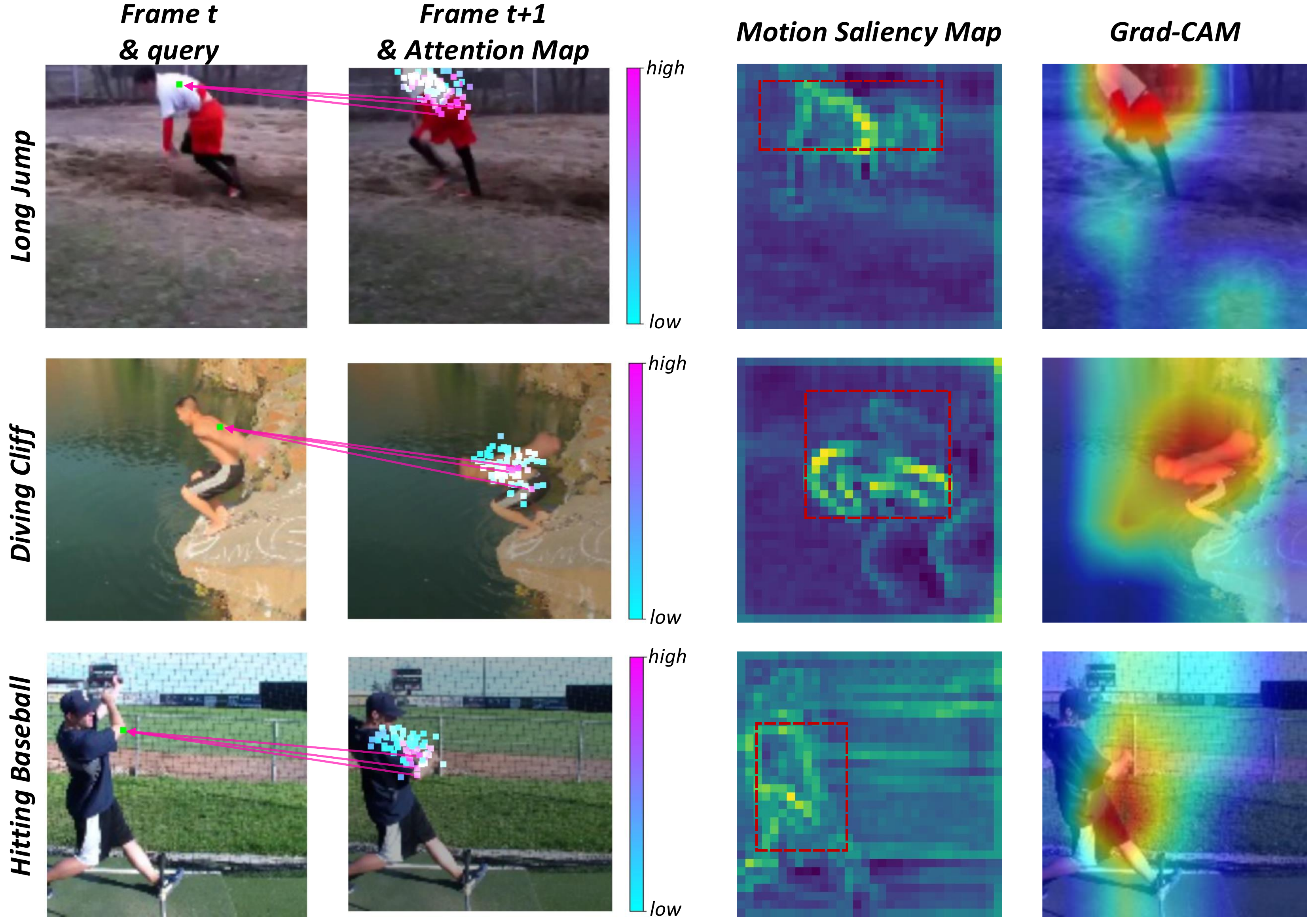}
\vspace{-0.13in}
\caption{\small Visualization of the inter-frame attention map, motion saliency map (MSM) and Grad-CAM \cite{Grad-CAM} of SIFA-Net for three videos in Kinetics-400. For the video in each row, the green point in its $t$-th frame denotes the query location. The correlation between query and sampling points in ($t$+1)-th frame (i.e., attention weight) is shown in heat map. We link the query and~sampling points with top-3 attention weights in purple line. The red box in MSM represents the region with highly salient object movements.}
\label{fig5:1}
\vspace{-0.27in}
\end{figure}

\subsection{Visualization Analysis of SIFA}
To better qualitatively examine SIFA block for video representation learning, we further visualize the inter-frame attention map over the locally deformable region, motion saliency map (MSM) and the class activation map with Grad-CAM \cite{Grad-CAM} of SIFA-Net (R50 backbone) in Figure \ref{fig5:1}. Note that we take the spatial location in center frame ($t$-th frame) of each sampled clip as the query, and employ the attention map of SIFA block in $res_5$ stage for visualization. In addition, for SIFA block with $k$=3, the deformable feature re-sampling is performed at two levels as in deformable ConvNets, leading to $9^2=81$ sampling points in ($t$+1)-th frame. As shown in the figure, the calculated motion saliency map of ($t$+1)-th frame matches the class activation map in general, which shows that the learnt MSM is able to capture the meaningful motion cues that benefit action classification. Through re-scaling deformable feature re-sampling with MSM, the sampling points are nicely adjusted according to the objects' scale, irregular shape, and large movements. This again confirms that SIFA block takes the object movement and deformation across frames into account to strengthen inter-frame feature alignment, thereby boosting temporal modeling.

\section{Conclusions and Discussions}
We have presented Stand-alone Inter-Frame Attention (SIFA) block, which explores the deformation across frames for temporal modeling with local self-attention.
Specifically, by taking the spatial location in current frame as query, SIFA performs self-attention over the keys/values in a local neighboring region of next frame.
Moreover, to tackle the irregular object deformation in next frame, a deformable design is leveraged to estimate the offset of each spatial location in local region, yielding the keys/values re-sampled in a deformation. Such deformable feature re-sampling is additionally re-scaled by motion cues to facilitate inter-frame attention learning. Finally, all deformable values are aggregated with attention to enhance per-frame feature. By plugging SIFA block into ResNet and Swin Transformer, we construct two new video backbones (SIFA-Net and SIFA-Transformer), and the experiments on four action recognition datasets demonstrate their effectiveness.

\textbf{Broader Impact.} One negative impact of this research in video representation learning is the significant environmental impact associated with training Transformer backbones, which are large and computationally expensive. There is also potential for these action recognition models to be misused, such as for unauthorized surveillance.

\textbf{Acknowledgments.} This work was supported by the National Key R\&D Program of China under Grant No. 2020AAA0108600.

{\small
\bibliographystyle{ieee_fullname}
\bibliography{egbib}

\begin{thebibliography}{10}\itemsep=-1pt

\bibitem{ViViT}
Anurag Arnab, Mostafa Dehghani, Georg Heigold, Chen Sun, Mario Lucic, and
  Cordelia Schmid.
\newblock {ViViT: A Video Vision Transformer}.
\newblock In {\em ICCV}, 2021.

\bibitem{Bertasius:ICML21}
Gedas Bertasius, Heng Wang, and Lorenzo Torresani.
\newblock {Is Space-Time Attention All You Need for Video Understanding?}
\newblock In {\em ICML}, 2021.

\bibitem{Carreira:CVPR17}
Joao Carreira and Andrew Zisserman.
\newblock {Quo Vadis, Action Recognition? A New Model and the Kinetics
  Dataset}.
\newblock In {\em CVPR}, 2017.

\bibitem{Chen:ECCV18}
Yunpeng Chen, Yannis Kalantidis, Jianshu Li, Shuicheng Yan, and Jiashi Feng.
\newblock {Multi-Fiber Networks for Video Recognition}.
\newblock In {\em ECCV}, 2018.

\bibitem{Dai:ICCV17}
Jifeng Dai, Haozhi Qi, Yuwen Xiong, Yi Li, Guodong Zhang, Han Hu, and Yichen
  Wei.
\newblock {Deformable Convolutional Networks}.
\newblock In {\em ICCV}, 2017.

\bibitem{Diba:CVPR17}
Ali Diba, Vivek Sharma, and Luc~Van Gool.
\newblock {Deep Temporal Linear Encoding Networks}.
\newblock In {\em CVPR}, 2017.

\bibitem{ViT}
Alexey Dosovitskiy, Lucas Beyer, Alexander Kolesnikov, Dirk Weissenborn,
  Xiaohua Zhai, Thomas Unterthiner, Mostafa Dehghani, Matthias Minderer, Georg
  Heigold, Sylvain Gelly, Jakob Uszkoreit, and Neil Houlsby.
\newblock {An Image is Worth 16x16 Words: Transformers for Image Recognition at
  Scale}.
\newblock In {\em ICLR}, 2021.

\bibitem{Fan:MVIT}
Haoqi Fan, Bo Xiong, Karttikeya Mangalam, Yanghao Li, Zhicheng Yan, Jitendra
  Malik, and Christoph Feichtenhofer.
\newblock {Multiscale Vision Transformers}.
\newblock {\em arXiv preprint arXiv:2104.11227}, 2021.

\bibitem{Fan:NIPS19}
Quanfu Fan, Chun-Fu Chen, Hilde Kuehne, Marco Pistoia, and David Cox.
\newblock {More Is Less: Learning Efficient Video Representations by Big-Little
  Network and Depthwise Temporal Aggregation}.
\newblock In {\em NeurIPS}, 2019.

\bibitem{Feichtenhofer:CVPR20}
Christoph Feichtenhofer.
\newblock {X3D: Expanding Architectures for Efficient Video Recognition}.
\newblock In {\em CVPR}, 2020.

\bibitem{Christoph:ICCV19}
Christoph Feichtenhofer, Haoqi Fan, Jitendra Malik, and Kaiming He.
\newblock {SlowFast Networks for Video Recognition}.
\newblock In {\em ICCV}, 2019.

\bibitem{Feichtenhofer:CVPR17}
Christoph Feichtenhofer, Axel Pinz, and Richard~P. Wildes.
\newblock {Spatiotemporal Multiplier Networks for Video Action Recognition}.
\newblock In {\em CVPR}, 2017.

\bibitem{Feichtenhofer:CVPR16}
Christoph Feichtenhofer, Axel Pinz, and Andrew Zisserman.
\newblock {Convolutional Two-Stream Network Fusion for Video Action
  Recognition}.
\newblock In {\em CVPR}, 2016.

\bibitem{Fischer:ICCV15}
Philipp Fischer, Alexey Dosovitskiy, Eddy Ilg, Philip Hausser, Caner Hazirbas,
  Vladimir Golkov, Patrick van~der Smagt, Daniel Cremers, and Thomas Brox.
\newblock {FlowNet: Learning Optical Flow with Convolutional Networks}.
\newblock In {\em ICCV}, 2015.

\bibitem{Kinetics:600}
Bernard Ghanem, Juan~Carlos Niebles, Cees Snoek, Fabian~Caba Heilbron, Humam
  Alwassel, Victor Escorcia, Ranjay Krishna, Shyamal Buch, and Cuong~Duc Dao.
\newblock {The ActivityNet Large-Scale Activity Recognition Challenge 2018
  Summary}.
\newblock {\em arXiv preprint arXiv:1808.03766}, 2018.

\bibitem{Goyal:SS}
Raghav Goyal, Samira~Ebrahimi Kahou, Vincent Michalski, Joanna Materzynska,
  Susanne Westphal, Heuna Kim, Valentin Haenel, Ingo Fruend, Peter Yianilos,
  Moritz Mueller-Freitag, Florian Hoppe, Christian Thurau, Ingo Bax, and Roland
  Memisevic.
\newblock {The "something something" video database for learning and evaluating
  visual common sense}.
\newblock In {\em ICCV}, 2017.

\bibitem{Kaiming:CVPR16}
Kaiming He, Xiangyu Zhang, Shaoqing Ren, and Jian Sun.
\newblock {Deep Residual Learning for Image Recognition}.
\newblock In {\em CVPR}, 2016.

\bibitem{Ji:PAMI}
Shuiwang Ji, Wei Xu, Ming Yang, and Kai Yu.
\newblock {3D Convolutional Neural Networks for Human Action Recognition}.
\newblock {\em IEEE Trans. on PAMI}, 2013.

\bibitem{Jiang:ICCV19}
Boyuan Jiang, MengMeng Wang, Weihao Gan, Wei Wu, and Junjie Yan.
\newblock {STM: SpatioTemporal and Motion Encoding for Action Recognition}.
\newblock In {\em ICCV}, 2019.

\bibitem{Sports1M}
Andrej Karpathy, George Toderici, Sanketh Shetty, Thomas Leung, Rahul
  Sukthankar, and Li Fei-Fei.
\newblock {Large-scale Video Classification with Convolutional Neural
  Networks}.
\newblock In {\em CVPR}, 2014.

\bibitem{Klaser:BMVC08}
Alexander Klaser, Marcin Marszalek, and Cordelia Schmid.
\newblock {A Spatio-Temporal Descriptor based on 3D-Gradients}.
\newblock In {\em BMVC}, 2008.

\bibitem{Kwon:ECCV20}
Heeseung Kwon, Manjin Kim, Suha Kwak, and Minsu Cho.
\newblock {MotionSqueeze: Neural Motion Feature Learning for Video
  Understanding}.
\newblock In {\em ECCV}, 2020.

\bibitem{Laptev:IJCV05}
Ivan Laptev.
\newblock {On Space-Time Interest Points}.
\newblock {\em International Journal of Computer Vision}, 64(2-3):107--123,
  2005.

\bibitem{Laptev:CVPR08}
Ivan Laptev, Marcin Marszalek, Cordelia Schmid, and Benjamin Rozenfeld.
\newblock {Learning Realistic Human Actions from Movies}.
\newblock In {\em CVPR}, 2008.

\bibitem{Dong:CVPR21}
Dong Li, Zhaofan Qiu, Yingwei Pan, Ting Yao, Houqiang Li, and Tao Mei.
\newblock {Representing Videos as Discriminative Sub-graphs for Action
  Recognition}.
\newblock In {\em CVPR}, 2021.

\bibitem{Dong:MM19}
Dong Li, Ting Yao, Zhaofan Qiu, Houqiang Li, and Tao Mei.
\newblock {Long Short-Term Relation Networks for Video Action Detection}.
\newblock In {\em ACM MM}, 2019.

\bibitem{Li:CVPR20}
Xianhang Li, Yali Wang, Zhipeng Zhou, and Yu Qiao.
\newblock {SmallBigNet: Integrating Core and Contextual Views for Video
  Classification}.
\newblock In {\em CVPR}, 2020.

\bibitem{Yan:CVPR20}
Yan Li, Bin Ji, Xintian Shi, Jianguo Zhang, Bin Kang, and Limin Wang.
\newblock {TEA: Temporal Excitation and Aggregation for Action Recognition}.
\newblock In {\em CVPR}, 2020.

\bibitem{Li:PAMI}
Yehao Li, Ting Yao, Yingwei Pan, and Tao Mei.
\newblock {Contextual Transformer Networks for Visual Recognition}.
\newblock {\em IEEE Trans. on PAMI}, 2022.

\bibitem{JiLin:ICCV19}
Ji Lin, Chuang Gan, and Song Han.
\newblock {TSM: Temporal Shift Module for Efficient Video Understanding}.
\newblock In {\em ICCV}, 2019.

\bibitem{Xingyu:CVPR19}
Xingyu Liu, Joon-Young Lee, and Hailin Jin.
\newblock {Learning Video Representations from Correspondence Proposals}.
\newblock In {\em CVPR}, 2019.

\bibitem{Swin-ViT}
Ze Liu, Yutong Lin, Yue Cao, Han Hu, Yixuan Wei, Zheng Zhang, Stephen Lin, and
  Baining Guo.
\newblock {Swin Transformer: Hierarchical Vision Transformer using Shifted
  Windows}.
\newblock In {\em ICCV}, 2021.

\bibitem{Liu:AAAI20}
Zhaoyang Liu, Donghao Luo, Yabiao Wang, Limin Wang, Ying Tai, Chengjie Wang,
  Jilin Li, Feiyue Huang, and Tong Lu.
\newblock {TEINet: Towards an Efficient Architecture for Video Recognition}.
\newblock In {\em AAAI}, 2020.

\bibitem{Liu:V-Swin}
Ze Liu, Jia Ning, Yue Cao, Yixuan Wei, Zheng Zhang, Stephen Lin, and Han Hu.
\newblock {Video Swin Transformer}.
\newblock {\em arXiv preprint arXiv:2106.13230}, 2021.

\bibitem{Long:CVPR19}
Fuchen Long, Ting Yao, Zhaofan Qiu, Xinmei Tian, Jiebo Luo, and Tao Mei.
\newblock {Gaussian Temporal Awareness Networks for Action Localization}.
\newblock In {\em CVPR}, 2019.

\bibitem{Long:ECCV20}
Fuchen Long, Ting Yao, Zhaofan Qiu, Xinmei Tian, Jiebo Luo, and Tao Mei.
\newblock {Learning to Localize Actions from Moments}.
\newblock In {\em ECCV}, 2020.

\bibitem{Long:TMM20}
Fuchen Long, Ting Yao, Zhaofan Qiu, Xinmei Tian, Tao Mei, and Jiebo Luo.
\newblock {Coarse-to-Fine Localization of Temporal Action Proposals}.
\newblock {\em IEEE Trans. on Multimedia}, 22(6):1577 -- 1590, 2020.

\bibitem{Loshchilov:ICLR17}
Ilya Loshchilov and Frank Hutter.
\newblock {SGDR: Stochastic Gradient Descent with Warm Restarts}.
\newblock In {\em ICLR}, 2017.

\bibitem{Luo:ICCV19}
Chenxu Luo and Alan Yuille.
\newblock {Grouped Spatial-Temporal Aggregation for Efficient Action
  Recognition}.
\newblock In {\em ICCV}, 2019.

\bibitem{Yue-Hei:CVPR15}
Joe Yue-Hei Ng, Matthew Hausknecht, Sudheendra Vijayanarasimhan, Oriol Vinyals,
  Rajat Monga, and George Toderici.
\newblock {Beyond Short Snippets: Deep Networks for Video Classification}.
\newblock In {\em CVPR}, 2015.

\bibitem{Qiu:ICCV17}
Zhaofan Qiu, Ting Yao, and Tao Mei.
\newblock {Learning Spatio-Temporal Representation with Pseudo-3D Residual
  Networks}.
\newblock In {\em ICCV}, 2017.

\bibitem{Qiu:ICML21}
Zhaofan Qiu, Ting Yao, Chong-Wah Ngo, and Tao Mei.
\newblock {Optimization Planning for 3D ConvNets}.
\newblock In {\em ICML}, 2021.

\bibitem{Qiu:CVPR19}
Zhaofan Qiu, Ting Yao, Chong-Wah Ngo, Xinmei Tian, and Tao Mei.
\newblock {Learning Spatio-Temporal Representation with Local and Global
  Diffusion}.
\newblock In {\em CVPR}, 2019.

\bibitem{Rocco:CVPR17}
Ignacio Rocco, Relja Arandjelovic, and Josef Sivic.
\newblock {Convolutional Neural Network Architecture for Geometric Matching}.
\newblock In {\em CVPR}, 2017.

\bibitem{Scovanner:MM07}
Paul Scovanner, Saad Ali, and Mubarak Shah.
\newblock {A 3-Dimensional SIFT Descriptor and Its Application to Action
  Recognition}.
\newblock In {\em ACM MM}, 2007.

\bibitem{Grad-CAM}
Ramprasaath~R. Selvaraju, Michael Cogswell, Abhishek Das, Ramakrishna Vedantam,
  Devi Parikh, and Dhruv Batra.
\newblock {Grad-CAM: Visual Explanations From Deep Networks via Gradient-Based
  Localization}.
\newblock In {\em ICCV}, 2017.

\bibitem{Simonyan:NIPS14}
Karen Simonyan and Andrew Zisserman.
\newblock {Two-stream Convolutional Networks for Action Recognition in Videos}.
\newblock In {\em NIPS}, 2014.

\bibitem{Srivastava:ICML15}
Nitish Srivastava, Elman Mansimov, and Ruslan Salakhutdinov.
\newblock {Unsupervised Learning of Video Representations using LSTMs}.
\newblock In {\em ICML}, 2015.

\bibitem{Sudhakaran:CVPR20}
Swathikiran Sudhakaran, Sergio Escalera, and Oswald Lanz.
\newblock {Gate-Shift Networks for Video Action Recognition}.
\newblock In {\em CVPR}, 2020.

\bibitem{Tran:ICCV15}
Du Tran, Lubomir Bourdev, Rob Fergus, Lorenzo Torresani, and Manohar Paluri.
\newblock {Learning Spatiotemporal Features with 3D Convolutional Networks}.
\newblock In {\em ICCV}, 2015.

\bibitem{Tran:ICCV19}
Du Tran, Heng Wang, Lorenzo Torresani, and Matt Feiszli.
\newblock {Video Classification with Channel-Separated Convolutional Networks}.
\newblock In {\em ICCV}, 2019.

\bibitem{Tran:CVPR18}
Du Tran, Heng Wang, Lorenzo Torresani, Jamie Ray, Yann LeCun, and Manohar
  Paluri.
\newblock {A Closer Look at Spatiotemporal Convolutions for Action
  Recognition}.
\newblock In {\em CVPR}, 2018.

\bibitem{Vaswani:NIPS17}
Ashish Vaswani, Noam Shazeer, Niki Parmar, Jakob Uszkoreit, Llion Jones,
  Aidan~N. Gomez, Lukasz Kaiser, and Illia Polosukhin.
\newblock {Attention Is All You Need.}
\newblock In {\em NIPS}, 2017.

\bibitem{Wang:CVPR11}
Heng Wang, Alexander Klaser, Cordelia Schmid, and Cheng-Lin Liu.
\newblock {Action Recognition by Dense Trajectories}.
\newblock In {\em CVPR}, 2011.

\bibitem{Wang:CVPR20}
Heng Wang, Du Tran, Lorenzo Torresani, and Matt Feiszli.
\newblock {Video Modeling with Correlation Networks}.
\newblock In {\em CVPR}, 2020.

\bibitem{Wang:CVPR21}
Limin Wang, Zhan Tong, Bin Ji, and Gangshan Wu.
\newblock {TDN: Temporal Difference Networks for Efficient Action Recognition}.
\newblock In {\em CVPR}, 2021.

\bibitem{Wang:ECCV16}
Limin Wang, Yuanjun Xiong, Zhe Wang, Yu Qiao, Dahua Lin, Xiaoou Tang, and
  Luc~Van Gool.
\newblock {Temporal Segment Networks: Towards Good Practices for Deep Action
  Recognition}.
\newblock In {\em ECCV}, 2016.

\bibitem{Wang:CVPR18}
Xiaolong Wang, Ross Girshick, Abhinav Gupta, and Kaiming He.
\newblock {Non-local Neural Networks}.
\newblock In {\em CVPR}, 2018.

\bibitem{Wang:ECCV18}
Xiaolong Wang and Abhinav Gupta.
\newblock {Videos as Space-Time Region Graphs}.
\newblock In {\em ECCV}, 2018.

\bibitem{WangAct:CVPR21}
Zhengwei Wang, Qi She, and Aljosa Smolic.
\newblock {ACTION-Net: Multipath Excitation for Action Recognition}.
\newblock In {\em CVPR}, 2021.

\bibitem{Philippe:ICCV13}
Philippe Weinzaepfel, Jerome Revaud, Zaid Harchaoui, and Cordelia Schmid.
\newblock {DeepFlow: Large Displacement Optical Flow with Deep Matching}.
\newblock In {\em ICCV}, 2013.

\bibitem{Xie:ECCV18}
Saining Xie, Chen Sun, Jonathan Huang, Zhuowen Tu, and Kevin Murphy.
\newblock {Rethinking Spatiotemporal Feature Learning: Speed-Accuracy
  Trade-offs in Video Classification}.
\newblock In {\em ECCV}, 2018.

\bibitem{Yao:AAAI21}
Ting Yao, Yiheng Zhang, Zhaofan Qiu, Yingwei Pan, and Tao Mei.
\newblock {SeCo: Exploring Sequence Supervision for Unsupervised Representation
  Learning}.
\newblock In {\em AAAI}, 2021.

\bibitem{Zhao:NIPS18}
Yue Zhao, Yuanjun Xiong, and Dahua Lin.
\newblock {Trajectory Convolution for Action Recognition}.
\newblock In {\em NeurIPS}, 2018.

\bibitem{Zhi:ICCV21}
Yuan Zhi, Zhan Tong, Limin Wang, and Gangshan Wu.
\newblock {MGSampler: An Explainable Sampling Strategy for Video Action
  Recognition}.
\newblock In {\em ICCV}, 2021.

\end{thebibliography}
}

\end{document}